\documentclass[10pt,twocolumn,letterpaper]{article}

\usepackage[pagenumbers]{format/cvpr} 

\newcommand{\ignore}[1]{}

\DeclareMathAlphabet{\mathbfit}{OML}{cmm}{b}{it}

\makeatletter
\DeclareRobustCommand\onedot{\futurelet\@let@token\@onedot}
\def\@onedot{\ifx\@let@token.\else.\null\fi\xspace}

\makeatother

\definecolor{MyDarkBlue}{rgb}{0,0.08,1}
\definecolor{MyAqua}{rgb}{0,0.7,0.7}
\definecolor{MyDarkGreen}{rgb}{0.02,0.6,0.02}
\definecolor{MyDarkOrange}{rgb}{0.40,0.2,0.02}
\definecolor{MyPurple}{HTML}{7338a0}
\definecolor{MyPink}{RGB}{250, 147, 240}
\definecolor{MyRed}{RGB}{227, 44, 94}
\definecolor{MyBlue}{RGB}{69, 144, 201}
\definecolor{MyOrange}{HTML}{d6970e}
\definecolor{MyPink}{HTML}{D86ECC}
\definecolor{MyGold}{rgb}{0.75,0.6,0.12}
\definecolor{MyDarkgray}{rgb}{0.66, 0.66, 0.66}

\newcommand{\IfDefinedSwitch}[3]{%
  \ifdefined#1
    #2 
  \else
    #3 
  \fi
}

\newcommand{\projectpage}{\href{https://red-fairy.github.io/ShadowDraw/}{project page}}
\newcommand{\shadowstroke}{shadow contour\xspace}
\newcommand{\shadowstrokes}{shadow contours\xspace}

\usepackage{pifont}
\usepackage{tabularx}

\usepackage{microtype}

\usepackage{amsmath,amsfonts,amsthm,amssymb}
\usepackage{mathtools}
\usepackage{bm}
\usepackage{nicefrac}
\usepackage{animate}

\usepackage{xcolor}
\usepackage{epsfig}
\usepackage{graphicx}

\usepackage{adjustbox}
\usepackage{array}
\newcolumntype{C}[1]{>{\centering\arraybackslash}p{#1}}
\usepackage{booktabs}
\usepackage{colortbl}
\usepackage{wrapfig}
\usepackage{hhline}
\usepackage{multirow}
\usepackage{subcaption}
\DeclareCaptionFont{between}{\fontsize{8pt}{9pt}\selectfont}
\usepackage{changepage}
\usepackage{extramarks}
\usepackage{fancyhdr}
\usepackage{lastpage}
\usepackage{setspace}
\usepackage{soul}
\usepackage{xspace}
\usepackage{adjustbox}

\usepackage{algpseudocode}
\usepackage{algorithmicx}
\usepackage[ruled]{algorithm2e}
\usepackage{enumitem}  
\usepackage{makecell}
\usepackage{pifont}
\usepackage{fontawesome}
\usepackage{ifthen}
\usepackage{etoolbox}
\usepackage{bbding}
\usepackage{pifont}

\usepackage{listings}
\usepackage{xcolor}

\definecolor{cvprblue}{rgb}{0.21,0.49,0.74}
\usepackage[pagebackref,breaklinks,colorlinks,allcolors=cvprblue]{hyperref}

\def\paperID{5615} 
\def\confName{CVPR}
\def\confYear{2026}

\title{ShadowDraw: From Any Object to Shadow–Drawing Compositional Art}

\author{
  \begin{tabular}{ccccc}
    Rundong Luo & Noah Snavely & Wei-Chiu Ma
  \end{tabular} \\ [2ex]
  Cornell University
}

\begin{document}

\newcommand{\arxiv}{arxiv version}
\newcommand{\embedVideo}{embed video}

\twocolumn[{%
\renewcommand\twocolumn[1][]{#1}%
\maketitle
\vspace{-10mm}
\captionsetup{type=figure}
\begin{center}
     \IfDefinedSwitch{\embedVideo}{\animategraphics[autoplay,loop,controls={play,stop}, width=0.8\linewidth, trim=0 0.2cm 0 0cm, clip]{30}{figures/teaser-cvpr/}{000}{149}
        }
        {\includegraphics[width=0.8\linewidth]{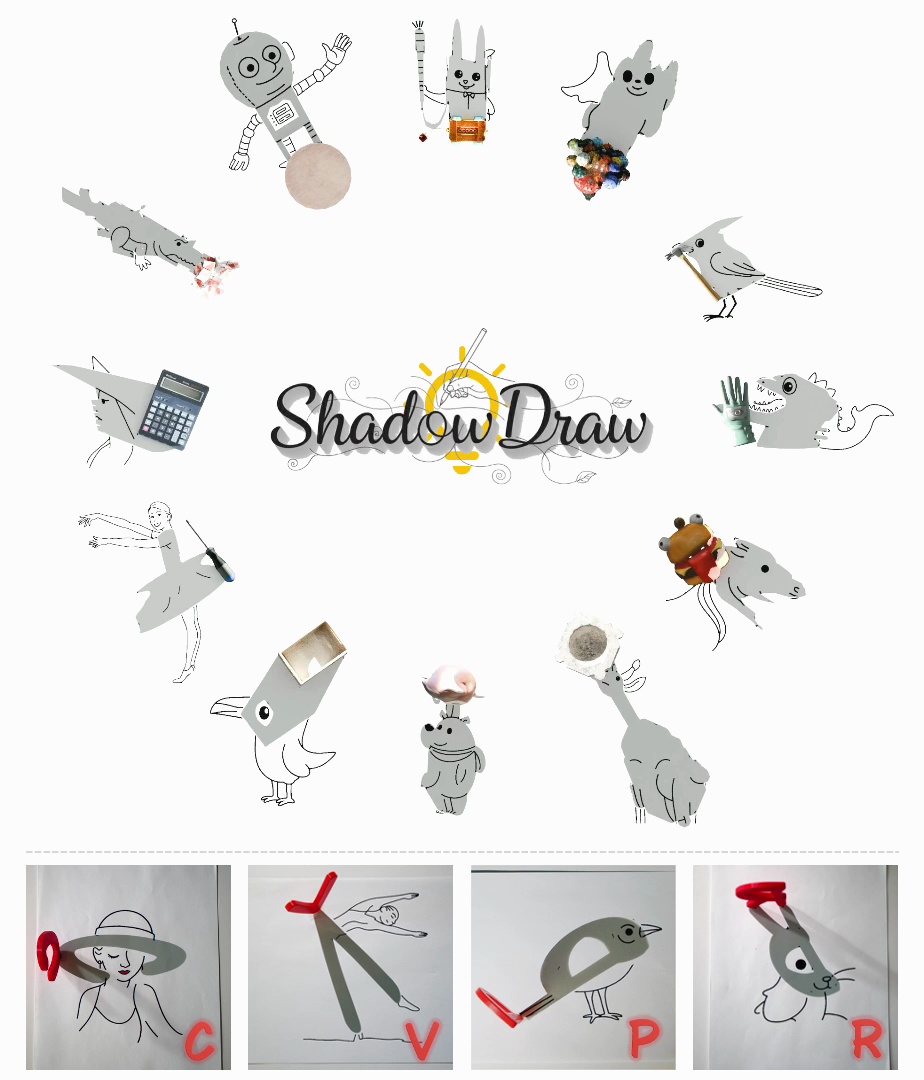}
        }
\end{center}
\vspace{-6mm}
\captionof{figure}{
    \textbf{Generating shadow–drawing compositional art.} Given an arbitrary 3D object, our framework jointly predicts scene parameters, including object pose and lighting, and generates a partial line drawing, such that the cast shadow seamlessly completes the drawing into a coherent image. 
    The system unites physical shadows with generative drawing, creating compelling compositions from cast shadows that provide only minimal structural cues.
    Our approach enables straightforward real-world deployment, as demonstrated with physical prototypes of letters C, V, P, R. 
    \emph{Best viewed in Adobe Acrobat Reader for the embedded animation.}
    }
\label{fig:teaser}
}
\vspace{1mm}
]

\begin{abstract}
We introduce \textsc{ShadowDraw}, a framework that transforms ordinary 3D objects into shadow–drawing compositional art. Given a 3D object, our system predicts scene parameters---including object pose and lighting---together with a partial line drawing, such that the cast shadow completes the drawing into a recognizable image. To this end, we optimize scene configurations to reveal meaningful shadows, employ \shadowstrokes to guide line drawing generation, and adopt automatic evaluation to enforce shadow-drawing coherence and visual quality. Experiments show that \textsc{ShadowDraw} produces compelling results across diverse inputs, from real-world scans and curated datasets to generative assets, and naturally extends to multi-object scenes, animations, and physical deployments. Our work provides a practical pipeline for creating shadow–drawing art and broadens the design space of computational visual art, bridging the gap between algorithmic design and artistic storytelling. 
 \IfDefinedSwitch{\arxiv}
 {Check out our~\projectpage~for more results and an end-to-end real-world demonstration of our pipeline!}
 {For more results and an end-to-end real-world demonstration of our pipeline, please refer to the \textbf{project page} in the supplementary material.}
\end{abstract}

\section{Introduction}

\begin{figure*}
    \centering
    \vspace{-3mm}
    \includegraphics[width=\linewidth,  trim=0 0 0.2cm 0.5cm]{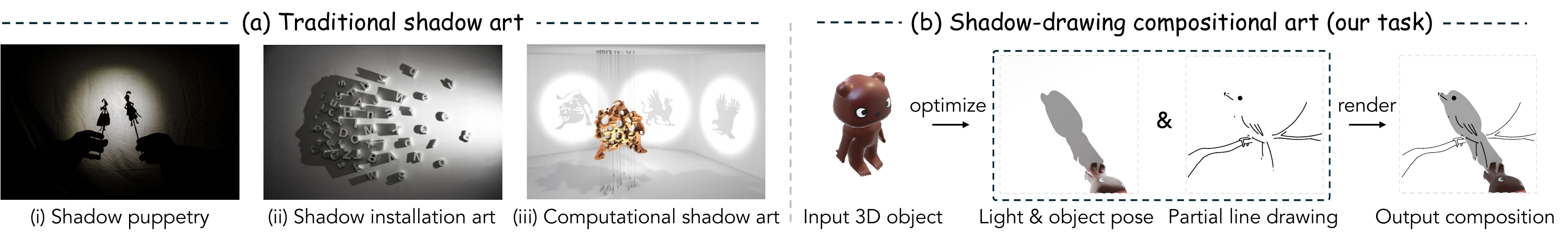}
    \vspace{-7.5mm}
    \caption{
    \textbf{From traditional shadow art to shadow–drawing compositional art.}  
    (a) Traditional shadow art, such as artist-crafted works and computational art designs, treats the objects and their cast shadows as the sole medium.
    (b) Our framework integrates shadows with line drawings: given a 3D object, we \emph{generate a partial drawing} and \emph{estimate scene parameters} (e.g., object pose and light position) such that the cast shadow completes the composition.}
    \vspace{-5mm}
    \label{fig:shadow-art}
\end{figure*}

Shadows have long captivated artists and audiences alike, serving as a powerful medium of expression across cultures and epochs. From traditional Chinese shadow theater, where cut-out puppets project intricate silhouettes, to contemporary shadow photography and installation art that manipulate light to craft evocative narratives (Fig.~\ref{fig:shadow-art}(a)(i–ii)), shadows transform absence into striking imagery. Defined by the shifting interplay of light and form, they are inherently ephemeral and visually expressive, embodying the delicate relationship between illumination, object, and perception.




Recent research in computational visual art has sought to formalize and extend shadow art practices. By framing the problem as an inverse design task, prior works optimize object geometry~\citep{mitra2009shadowart,sadekar2022shadowart}, material properties~\citep{baran2012layered,min2017softshadow}, and lighting~\citep{weyrich2009fabricating, pereira2014computational} to achieve desired visual effects (Fig.~\ref{fig:shadow-art}(a)(iii)). While effective, these approaches 
assume a predefined visual target 
(\emph{i.e.}, knowing a priori what to generate) and rely on parameter optimization to reproduce it.

In this paper, 
we are instead interested in exploring the 
interplay between physical shadows and generative drawing. 
Our work is directly inspired by Belgian artist Vincent Bal~\citep{VincentBalChannel},
whose playful works reveal how the cast shadows of everyday objects can seamlessly complete drawn elements. Motivated by this 
idea, we aim to develop a computational framework that 
captures
the same sense of serendipity: unified compositions where shadow and line drawing contribute to a cohesive whole, without relying on a predefined target.

Formally, given a 3D object as input, we aim to predict both scene parameters (light direction and object pose) and a partial line drawing such that, when the scene is illuminated, the shadow cast by the object completes the drawing into a coherent, recognizable image (Fig.~\ref{fig:shadow-art}(b)). 
However, achieving such compositions is extremely challenging. First, we do not know \emph{a priori} what subject to generate, yet generative models typically require detailed prompts to produce high-quality results. Second, effective image conditioning is essential, but rendered inputs such as the shadow image or object–shadow composite provides weak structural cues provides weak structural cues, often yielding generated compositions in which the shadow contributes little. This difficulty is further compounded by the scarcity of shadow–drawing examples, leaving only limited data for training.

To address these challenges, we reformulate the problem around the \emph{\shadowstroke}---the boundary contour of the shadow. Although a raw shadow and its contour encode the same geometry, we empirically find that reducing the shadow to a clean binary outline provides stronger conditioning, yielding tighter alignment between the shadow and drawing. The \shadowstroke~also naturally supports scalable data construction, as closed contours can be efficiently extracted from line drawings, and further enables the use of pretrained edge-conditioned generative models for line drawing generation. Building on this idea, we first train a line drawing generator conditioned jointly on text and the~\shadowstroke. Then we search over scene parameters by differentiable rendering to discover semantically interesting shadows and leverage VLMs to create detailed descriptions of the intended drawing from the~\shadowstrokes, guided by carefully designed in-context examples. Finally, we design automatic metrics to verify shadow-drawing coherence and visual quality, retaining only those compositions where the shadow meaningfully contributes.

Extensive experiments demonstrate the effectiveness of our framework in generating coherent and visually compelling shadow–drawing compositions. Our approach generalizes across diverse object models, including 3D datasets, real-world scans, and generated assets~(Fig.~\ref{fig:teaser}), and naturally extends to multi-object scenes, animated settings, and physical deployments~(Fig.~\ref{fig:application}). 
Notably, the required physical setup is simple: an everyday object (whose 3D model can be easily scanned using mobile app such as PolyCam), a planar surface, and a spotlight are sufficient to reproduce our computational designs. This accessibility broadens the expressive range of shadow art and significantly lowers the barrier for exploring this emerging form of visual storytelling.

\section{Related Work}
\label{sec:related-works}

\noindent\textbf{Computational Visual Art.}  
Computational visual art explores artistic creations whose perceptual impact depends on 3D geometry, material properties, and controlled illumination. Works span sculpture, architecture, and fabricated objects, often exploiting optical phenomena such as shadows, reflections, or refractions to produce striking imagery~\citep{wu2022survey}. 
Traditionally, the creation process relies heavily on an artist’s intuition and iterative trial-and-error, making it labor-intensive and technically demanding. 
Computational approaches aim to 
assist this process by formulating it as an inverse problem: given a target visual effect, optimize scene parameters---e.g., object pose, geometry, and light configuration---so that the rendered appearance matches the design intent. 
Techniques range from combinatorial optimization for occluder placement~\citep{min2017softshadow,baran2012layered} to differentiable rendering for shape refinement~\citep{wu2022mirrorcup,sadekar2022shadowart}. Applications span reflection-based art~\citep{weyrich2009fabricating}, volumetric displays~\citep{hirayama2019projection}, and multiview illusions~\citep{illusion3d}. 
Unlike these approaches, which optimize scene parameters to achieve a \emph{given} target visual effect, our system \emph{jointly} estimates both the scene parameters and the target subject, making the task much more challenging. 

\smallskip\noindent\textbf{Line Drawing Generation.}  
Line drawings have long been studied as a compact yet expressive representation of shape and semantics. 3D-based approaches derive drawings directly from geometry, extracting contours, depth cues, or neural features to approximate sketches~\citep{decarlo2003suggestive,judd2007apparent,liu2020neuralcontours,gryaditskaya2020lifting}. In parallel, image-based methods framed line drawing as a supervised translation task, mapping photographs to vector strokes or contours using paired~\citep{li2019photo,li2019im2pencil} or unpaired data~\citep{chan2022line}. Beyond image-to-drawing, recent research has explored diverse tasks, including text-guided generation~\citep{frans2022clipdraw}, sketch animation~\citep{jiang2012trajectory,bertasius2019learning}, object sketching~\citep{liu2021neural,li2017free}, portrait rendering~\citep{yi2019apdrawinggan}, line drawing completion~\citep{bhunia2022doodleformer,liu2019sketchgan}, and sequential or collaborative creation~\citep{vinker2025sketchagent}. Despite these advances, prior studies primarily regard line drawing as an isolated modality, whereas our work explicitly connects it with cast shadows to create hybrid visual compositions.

\smallskip
\noindent\textbf{Shadow Art.} Shadow art is a subclass of computational visual art where the shadow cast by a physical object under a controlled light source is used for artistic expression. 
Early computational methods focused on single-light, binary-shadow designs, deforming object geometry so that the resulting shadow matched the desired shape~\citep{mitra2009shadowart,hsiao2018multiviewwire}. With the advent of differentiable rendering, object shapes can be directly optimized with respect to the shadow image under more sophisticated setups~\citep{sadekar2022shadowart,qu2024wired}. 
However, these approaches may produce irregular or impractical geometries, limiting real-world applicability. Recent methods broaden the design space to include multi-layer occluder systems~\citep{min2017softshadow} and color shadow projections using translucent materials~\citep{baran2012layered}. Beyond rigid objects, shadow art has been explored using human bodies~\citep{won2016shadowtheatre} and hand gestures~\citep{hand-shadow-poser}. 
In contrast to prior work that treats shadows as the sole visual medium, we explore the interplay between such physical effects in 3D space and generative models operating in the pixel domain. Given a candidate cast shadow, which may offer only the barest suggestion of the underlying object, the system must imagine how to complete it into an interesting composition.

\section{Generative Shadow-Drawing Art}
\label{sec:method}

\begin{figure*}
    \vspace{-2.5mm}
    \centering
    \includegraphics[width=\linewidth, trim=0.5cm 1.5cm 0 0.5cm,]{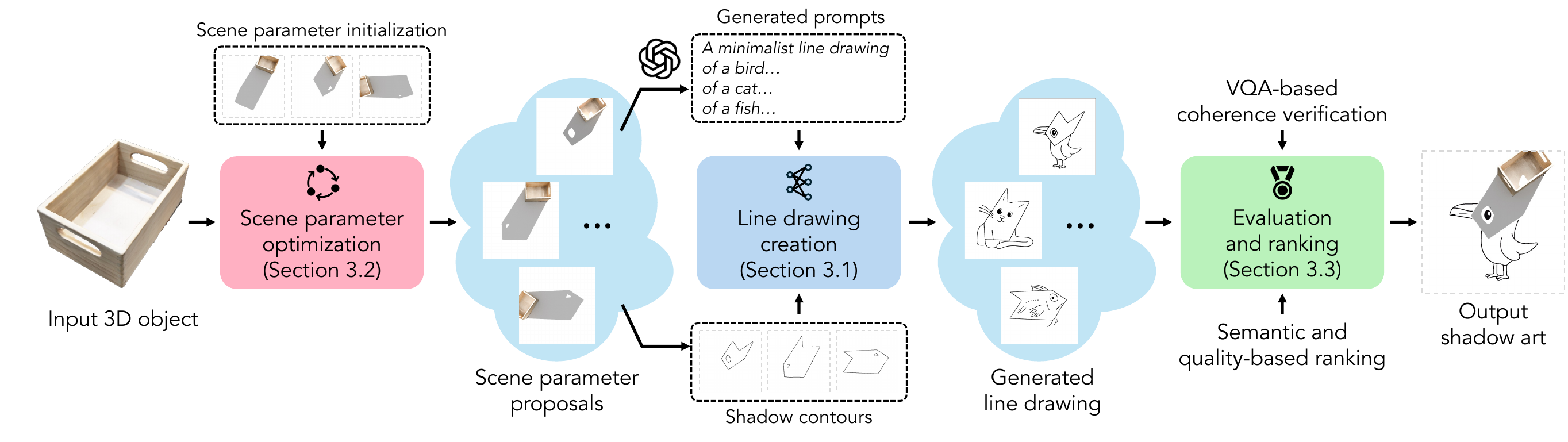}
    \captionsetup{font=small}
    \vspace{-3mm}
    \caption{\textbf{Framework overview.} Given a 3D object, we first optimize scene parameters specifying the object pose and light configuration. From the rendered shadows, we derive text prompts with VLM and extract \shadowstrokes, which together condition the line drawing generator. The generated drawings are then filtered using a VQA-based coherence check and ranked by semantic and quality metrics. The final output is a partial line drawing along with scene parameters that, when rendered, form a coherent shadow–drawing composition.}
    \captionsetup{font=normal}
    \label{fig:pipeline}
    \vspace{-4mm}
\end{figure*}

We explore an art form that unifies shadow and line drawing into a single, coherent composition. The input is a 3D object model, and the outputs are: (i) a partial line drawing, and (ii) scene parameters specifying the 3D position and direction of a light source and the pose of the object. These elements can be jointly arranged so that, when illuminated, the cast shadow completes the line drawing, producing a recognizable image on the projection surface.

In our setup, the canvas lies on the ground plane, and the light source is modeled as a spotlight to produce sharp, coherent shadows. The light maintains a fixed distance from the canvas center, yielding two degrees of freedom: elevation and azimuth. The object can rotate about its vertical axis and translate along two axes on the ground. We normalize the object’s longest dimension to standardize shadow size; in physical deployments, the canvas can be inversely scaled to preserve the ratio. The line drawing is rendered in black, and the shadow is represented as a gray silhouette, with the final composition emerging from their precise spatial alignment. For simplicity, we restrict our study to single-light configurations and omit surface textures.

\smallskip
\noindent\textbf{Overview.} Our task is highly under-constrained: a single 3D object admits countless combinations of light, object, and line drawing arrangements, and traditional practice relies on artistic intuition to achieve compelling results. To make the problem tractable, we first consider a simplified setting where the object pose and light direction are fixed, aiming to generate a line drawing that seamlessly integrates with the shadow. We then extend to optimizing scene parameters to produce visually distinctive and semantically meaningful shadows, accompanied by textual descriptions of the intended subject. Finally, we summarize the compositions by selecting a small set of high-quality examples. 

\subsection{From Shadow Contour to Shadow-Drawing Art}
\label{sec:generate-line-drawing}

We begin with a simplified setting where the scene parameters are fixed, the subject description is given, and the goal is to generate a line drawing that integrates seamlessly with the rendered shadow. 


A straightforward baseline is to train an image-conditioned generative model on rendered inputs such as the shadow image or the object–shadow composite. However, this approach faces two challenges: weak conditioning signals, which limit the model’s ability to align the drawing with the shadow, and data scarcity, as only a few dozen shadow–drawing examples exist online.

To address these issues, we replace the raw shadow with its 2D boundary contour, referred to as the \emph{\shadowstroke}, as the conditioning input. Although a raw shadow and its contour encode the same geometry, reducing it to a clean binary outline provides stronger conditioning: models trained on grayscale shadows often drift from the intended geometry, whereas contour-based conditioning achieves tighter alignment. Beyond being a strong geometric cue, reformulating the task as transforming closed contours into line drawings yields two benefits: (i) it enables the use of well-established edge-conditioned generative models, and (ii) it allows scalable data synthesis, since shadow-like contours can be efficiently extracted from generic line drawings. As we will demonstrate in the experiments, this \shadowstroke~design significantly improves generation quality.

We first describe how we curate data for training (see supplementary~\ref{appendix:data-contruction} for details). First, we generate a set of line drawings using GPT-4o and retain only those containing regions bounded by strokes. We then train a FLUX-1-dev LoRA~\citep{flux} on this filtered set and use it to synthesize an additional 10K line drawings from GPT-4o-generated prompts about everyday subjects. Closed contours extracted from these drawings serve as \shadowstroke~conditions.

Using this dataset, we train a latent flow-based model $\boldsymbol{\epsilon}_\theta$ with the standard score-matching objective:
\begin{equation}
\min_{\theta} \mathbb{E}_{\mathbf{x}_0, \boldsymbol{\epsilon}, \mathbf{c}_i, \mathbf{c}_t, t} \left\| \omega(t) \left( \boldsymbol{\epsilon}_\theta(\mathbf{x}_t, \mathbf{c}_i, \mathbf{c}_t, t) - \boldsymbol{\epsilon} \right) \right\|^2,
\end{equation}
where $\mathbf{x}_0$ is the target line-drawing latent, $t$ is the timestep, 
$\mathbf{x}_t$ is the noisy sample at $t$, $\mathbf{c}_i$ is the \shadowstroke~condition, and $\mathbf{c}_t$ is the text prompt. We use FLUX.1-Canny~\citep{flux} as the base model and train a LoRA adapter~\citep{dora} on top.

At inference time, we render the shadow given the scene parameters, extract its boundary, and condition the model jointly on this contour and the text description. To prevent strokes from overlapping the object, we treat generation as an outpainting problem~\citep{repaint} with a binary object mask $\mathbf{m}$, preserving masked regions during denoising:
\begin{align}
\mathbf{x}_{t} &= \mathbf{m} \odot \mathbf{x}_{t}^{\text{mask}} + (1 - \mathbf{m}) \odot \hat{\mathbf{x}}_{t}, \\ \mathbf{x}_{t}^{\text{mask}} &\sim \mathcal{N}(\sqrt{\bar{\alpha}_t}x_0^{\text{mask}}, (1 - \bar{\alpha}_t)\mathbf{I}),
\end{align}
where $\mathbf{x}_{0}^{\text{mask}}$ is the latent of the mask $\mathbf{m}$ and $\hat{\mathbf{x}}_{t}$ is the model prediction at timestep $t$.

Finally, we erase input \shadowstroke~from the generated drawing, reinsert the 3D object, and render the composition in which the cast shadow completes the drawing.

\subsection{Scene Configuration Selection}
\label{sec:scene-config-select}

With the line drawing generation model in place, we relax the fixed-setting assumption and explore how to discover scene configurations that yield diverse and semantically meaningful shadows. For each candidate configuration, we create a text prompt describing the intended subject, which conditions the subsequent line drawing generation.

\smallskip
\noindent\textbf{Scene Parameter Optimization.} We consider five scene parameters: light azimuth $\theta$, elevation $\phi$, object center position in polar coordinates $(r, \gamma)$, 
and the object rotation about its vertical axis $\alpha$. To ensure the shadow extends toward the canvas center, we set $\gamma = \theta$ and $r = 0.8 \times$ the canvas radius, leaving three degrees of freedom. 

We quantify shadow quality using fractal dimension (FD)~\citep{falconer2013fractal}, which measures contour complexity via multi-granularity box counting; a higher FD indicates more irregular, visually rich shapes. To allow gradient-based optimization, we use a differentiable approximation of FD:
\begin{equation}
    \mathcal{L} = -\mathrm{FD}(\mathbf{S}), \quad \mathbf{S} = \mathrm{Renderer}(\theta, \phi, r, \gamma, \alpha),
\end{equation}
where $\mathbf{S}$ denotes the rendered binary shadow, obtained via PyTorch3D differentiable silhouette rendering. 

Empirically, we initialize the search with 48 configurations, spanning 12 azimuths in 30$^\circ$ increments and 4 elevations to vary shadow length, each combined with a random vertical object rotation to diversify shapes. For each initialization, parameter updates are restricted to its local neighborhood to prevent optimization space overlap of different initialization and ensure scene parameters remain physically plausible and easily reproducible in real-world setups.

\smallskip
\noindent\textbf{Visual Prompt Proposal.} Detailed prompts are known to substantially improve text-to-image generation~\citep{t2iprompts}, whereas generic descriptions such as ``a man'' or ``a bird'' are often too vague to produce coherent shadow–drawing compositions.  Unlike prior computational art methods that rely on a small set of hand-crafted prompts~\citep{illusion3d,visualanagram}, our goal is to support arbitrary inputs while adapting to diverse shadow geometries. This calls for an automated pipeline that generates scene-specific prompts directly from the shadow itself.

To achieve this, we employ vision–language models. The model is instructed to imagine a line drawing in which the given \shadowstroke~naturally serves as a key structural element, and to produce a detailed description of that drawing. For flexible control, users may specify the desired subject by modifying the prompt accordingly. To encourage reasoning about the stroke’s geometry and to maintain both semantic and visual coherence, we adopt a chain-of-thought–style prompting template, with the complete system prompt provided in the supplementary~\ref{appendix:visual-prompt-proposal}. 


\subsection{Evaluation and Ranking}
\label{sec:filtering}

Having generated compositions across the proposed scene configurations, a natural question arises: \emph{which results are worth keeping?} Not all samples are reliable—some exhibit poor alignment between the drawing and shadow, while others contain shadows that contribute little to the final image. To retain only strong outputs, we apply a systematic filtering process along three dimensions: (i) \emph{shadow–drawing coherence}, (ii) \emph{shadow’s contribution}, and (iii) \emph{visual quality}. These criteria ensure that the selected compositions are visually compelling and structurally coherent. Additional visual examples are provided in the supplementary~\ref{fig:ranking}.

\begin{figure}[t!]
  \centering
  \vspace{-3mm}
  \includegraphics[width=0.88\linewidth,trim=0.4cm 0cm 0cm 0,]{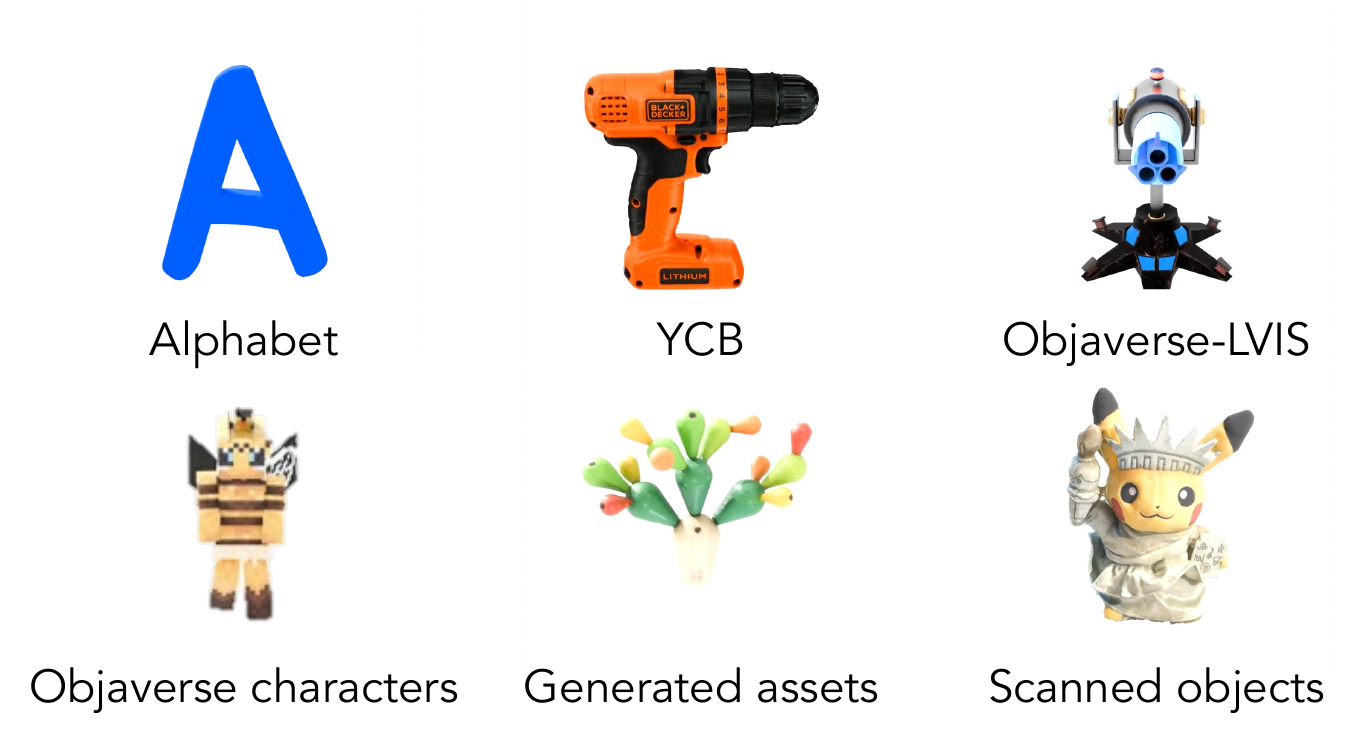}
    \vspace{-3.5mm}
  \captionsetup{font=small}
  \caption{Examples of objects for evaluation.}
  \captionsetup{font=normal}
  \vspace{-4.5mm}
  \label{fig:objects}
\end{figure}


\smallskip
\noindent\textbf{Shadow-drawing Coherence Verification.}  
We adopt a VQA-based verification strategy~\citep{vqascore} to measure the coherence between the given stroke and the line drawing. During the Visual Prompt Proposal stage, the VLM is instructed to specify the intended role of the \shadowstroke~in the composition (e.g., ``the body of a fish''). We then overlay the \shadowstroke~in red onto the generated line drawing and query another VLM with a yes/no question: \emph{``Does the highlighted stroke outline the described component?''} Candidates receiving a ``no'' response are discarded.

\smallskip
\noindent\textbf{Shadow Contribution Assessment.}  
We further evaluate whether the shadow meaningfully enhances the composition. Specifically, we compare the complete line drawing (\emph{full}) with a variant where the \shadowstroke~is removed (\emph{partial}), evaluating both with CLIP similarity~\citep{clip} and human-preference-aligned metrics, including ImageReward~\citep{imagereward} and Human Preference Score (HPS)~\citep{hps}. If the partial version attains a higher ImageReward or HPS score, the composition is discarded, indicating that the shadow fails to improve the overall generation quality.

\smallskip
\noindent\textbf{Ranking.}  
For the remaining candidates, we compute an improvement score for each metric:
\begin{align}
    \Delta_{\mathrm{CLIP}} &= \mathrm{CLIP}_{\mathrm{full}}^2  / \mathrm{CLIP}_{\mathrm{partial}}^2, \\
    \Delta_{\mathrm{IR}} &= \Phi(\mathrm{IR}_{\mathrm{full}})^2 - \Phi(\mathrm{IR}_{\mathrm{partial}})^2, \\
    \Delta_{\mathrm{HPS}} &= \mathrm{HPS}_{\mathrm{full}}^2 - \mathrm{HPS}_{\mathrm{partial}}^2,
\end{align}
where $\Phi(\cdot)$ denotes the CDF of the standard Gaussian, as ImageReward scores are normalized. The overall ranking score is defined as:
\begin{equation}
\mathcal{R} = \Delta_{\mathrm{CLIP}} \cdot \Delta_{\mathrm{IR}} \cdot \Delta_{\mathrm{HPS}},
\end{equation}
and top-$K$ compositions by $\mathcal{R}$ are selected as final outputs.

\section{Experiments}
\label{sec:experiments}

\begin{figure*}[t!]
    \centering
    \vspace{-3.5mm}
    \includegraphics[width=\linewidth,trim=0.5cm 0 0.5cm 0,]{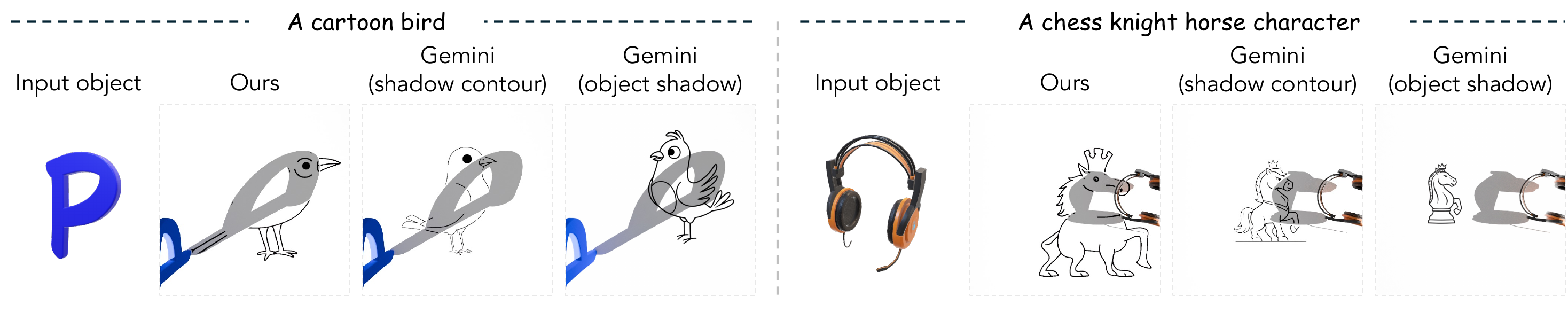}
    \vspace{-7.5mm}
    \captionsetup{font=small}
    \caption{\textbf{Qualitative baseline comparisons.} Large models like Gemini fail to capture the subtle notion of shadow–drawing art and often produce outputs where the shadow contributes little, whereas our method yields coherent shadow–drawing compositions of better quality.}
    \captionsetup{font=normal}
    \label{fig:baseline}
    \vspace{-3mm}
\end{figure*}

In this section, we first detail our experimental setup, then present quantitative comparisons with baselines and ablation variants, followed by qualitative results across diverse 3D assets. Finally, we demonstrate downstream applications that our framework enables out of the box.

\subsection{Experimental Setup}

\textbf{Baselines.} Since no existing method explicitly targets shadow–drawing compositional art, we construct two baselines using state-of-the-art image generation models. The first, \emph{Gemini (object–shadow)}, employs Gemini Flash 2.5 Image~\citep{gemini} to generate the composition conditioned on both the object–shadow composite and the text prompt produced by our approach. The second, \emph{Gemini (\shadowstroke)}, replaces the object-shadow composite with the \shadowstroke~image, providing more precise geometric guidance. Details of baseline execution are provided in supplementary material~\ref{appendix:baseline-execution}. To analyze the contribution of each framework component, we further conduct ablations by (i) training the line drawing model on artist-sourced images conditioned on the object–shadow composite, (ii) training the same model conditioned on \shadowstroke, and (iii) initializing scene parameters randomly without optimization.

\begin{figure*}
    \centering
    \vspace{-8mm}
    \includegraphics[width=0.88\linewidth]{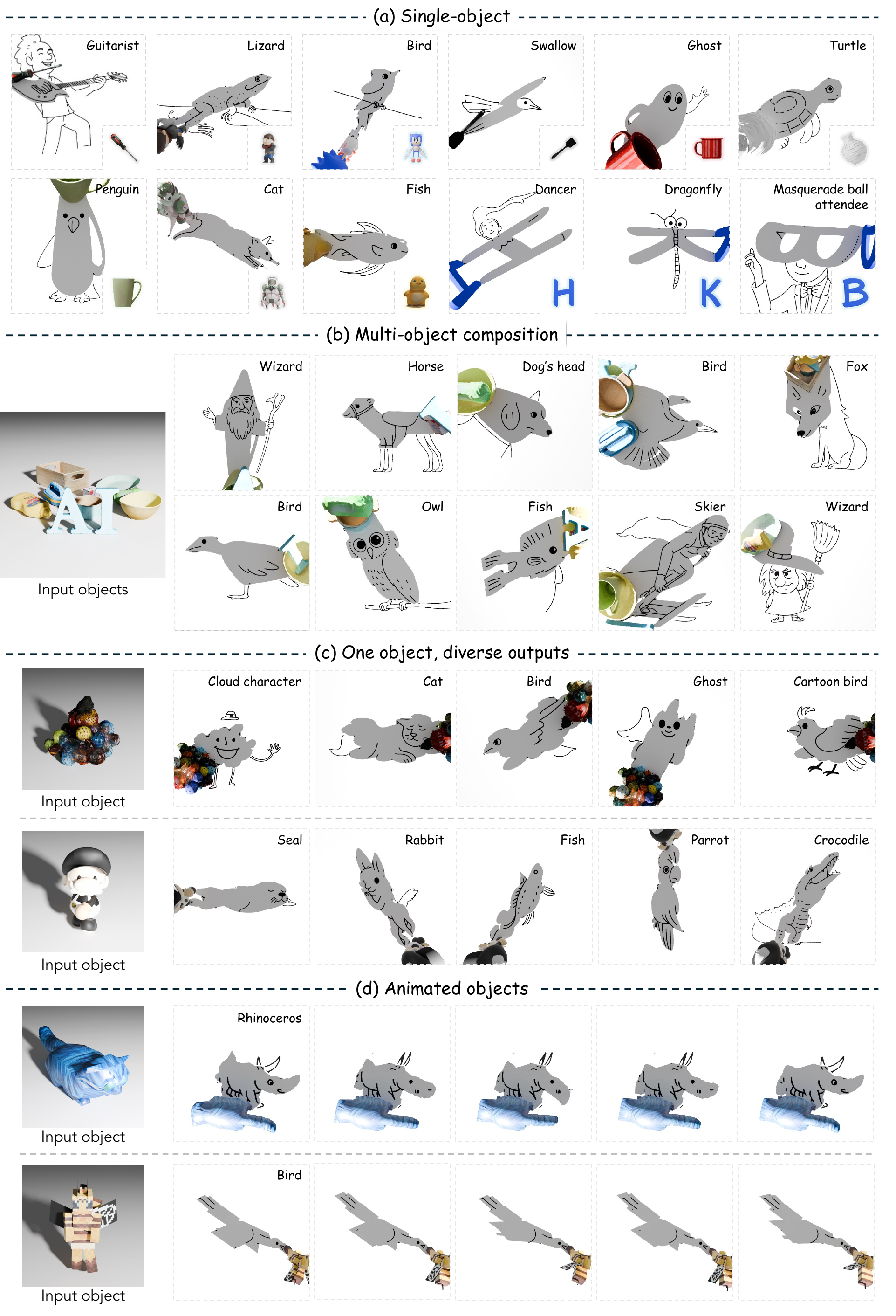}
    \captionsetup{font=small}
    \vspace{-3.5mm}
    \caption{\textbf{Gallery of shadow-drawing art generation results.} (a) Single-object generation. (b) Multi-object compositions. (c) Diverse results from the same object by varying light, pose, and line drawing. (d) Animated shadow–drawing art, where the shadow evolves with the motion of the object to complete the composition. More results on our \textbf{project page} in the supplementary material.}
    \captionsetup{font=normal}
    \label{fig:application}
\end{figure*}

\begin{table}[t]
  \centering\small
  \scalebox{0.84}{
  \begin{tabular}{l|ccc}
    \toprule
    Method   & CLIP$\uparrow$ & Conceal$\uparrow$ & Human Preference (\%)\\
    \midrule
    Gemini (object-shadow) & 31.28 & -0.2840 & 3.6\% \\
    Gemini (\shadowstroke) & 31.65 & 0.2421 & 6.0\% \\
    \midrule
    Ours & \textbf{32.41} & \textbf{3.0059} & \textbf{70.4\%} \\
    \bottomrule
  \end{tabular}
  }
  \vspace{-2mm}
  \captionsetup{font=small}
  \caption{Comparison with the baselines.}
  \captionsetup{font=normal}
  \vspace{-6mm}
  \label{tab:baseline}
\end{table}

\smallskip
\noindent\textbf{Data.}  
We collect 200 object models from diverse sources to evaluate our framework~(examples in Fig.~\ref{fig:objects}). Specifically, the dataset includes 26 alphabet models (A--Z), 20 from the YCB robotics dataset~\citep{ycb}, 87 objects from distinct categories in Objaverse-LVIS~\citep{objaverse}, and 30 character models from Objaverse. To demonstrate robustness to imperfect geometry, we further include 20 real-world household objects scanned via Polycam and 17 synthetic assets generated by MeshLRM~\citep{wei2024meshlrm}. For training ablation baselines without our synthetic data, we collect 71 images from the artist’s YouTube channel~\citep{VincentBalChannel}, manually filtered and processed to extract the object, shadow, and line-drawing components.

\smallskip
\noindent\textbf{Metrics.} For quantitative evaluation, we compare our method against the baselines among the top-4 ranked outputs. Specifically, we report the average CLIP score \citep{hessel2021clipscore}, ImageReward~\citep{imagereward}, and Human Preference Score \citep{hps} of the generated compositions. We further measure concealment~\citep{visualanagram}, defined as the CLIP score difference between the complete line drawing and the version with the \shadowstroke~removed. Additionally, we conduct two user studies. In the first, 10 participants compare the top-1 outputs from our method and the two baselines, selecting the preferred result or indicating no preference. In the second, 8 participants evaluate the success rate of our framework by selecting any number of satisfactory results among the top-4 images produced by our evaluation and ranking pipeline.

\subsection{Results and Analyses} 

\noindent\textbf{Baseline Comparisons.}  
Tab.~\ref{tab:baseline} compares our framework with the baselines. In the \emph{Gemini (object–shadow)} setting, the composite image provides weak structural cues, insufficient for forming coherent shadow–drawing compositions. The \shadowstroke~representation offers stronger geometric guidance yet remains limited, as pretrained image generators still struggle to capture the nuanced interplay between shadow and drawing. Consequently, both baselines yield low concealment scores, suggesting that the shadow contributes little or even negatively to the final composition.  
User studies align with these findings: participants favored our method in 70.4\% of cases, with 20.1\% marked as ``neither preferable.'' Our approach also achieves much higher concealment scores, confirming that the shadow serves as a crucial structural element rather than a redundant component. Qualitative results in Fig.~\ref{fig:baseline} further highlight our higher visual quality. 
Moreover, \textbf{96.8\%} of the top-4 results generated by our framework include at least one rated satisfactory by users, validating its overall effectiveness.

\begin{figure*}[t!]
    \centering
    \vspace{-2.5mm}
    \includegraphics[width=\linewidth,trim=1cm 1cm 0cm 0.2cm,]{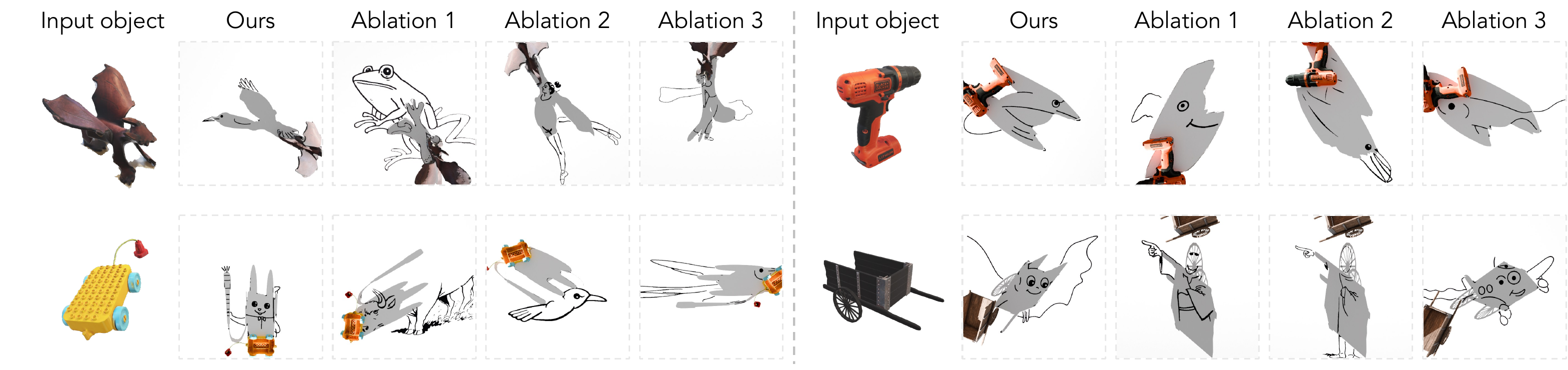}
    \vspace{-4mm}
    \captionsetup{font=small}
    \caption{\textbf{Qualitative ablation studies.} While ablated variants often yield results where the shadow contributes little, our method produces compositions with better visual quality and stronger shadow–drawing coherence.}
    \captionsetup{font=normal}
    \vspace{-4mm}
    \label{fig:ablation}
\end{figure*}

\smallskip
\noindent\textbf{Ablation Studies.}  
We conduct ablation studies to evaluate the contributions of three key components: the proposed \shadowstroke condition, the synthetic training dataset, and scene parameter optimization. As shown in Tab.~\ref{tab:ablation}, each component yields clear gains. Replacing object–shadow conditioning with \shadowstroke notably improves both generation quality and concealment; substituting the limited artist-sourced data with our large-scale synthetic dataset further boosts all metrics; and incorporating scene parameter optimization delivers the strongest overall performance. Qualitative comparisons in Fig.~\ref{fig:ablation} illustrate these effects. We further justify the effectiveness of our evaluation algorithm with user studies. See analyses in supplementary~\ref{appendix:ranking}.

\begin{table*}[t]
    \vspace{-1mm}
    \centering\small
    \scalebox{0.9}{
    \begin{tabular}{l|C{22mm}C{22mm}C{26mm}|C{12mm}C{12mm}C{12mm}C{12mm}}
    \toprule
        Method & Condition type & Training data & \makecell{Scene param optim.} & CLIP~$\uparrow$ & Conceal~$\uparrow$ & IR~$\uparrow$ & HPS~$\uparrow$\\
    \midrule
       Ablation 1  & object-shadow & artist source & $\checkmark$ & 31.04 & 0.225 & -0.0720 & 0.2244 \\
       Ablation 2  & \shadowstroke~& artist source & $\checkmark$ & 31.38 & 2.215 & 0.1552 & 0.2269 \\
       Ablation 3  & \shadowstroke~& synthetic & $\times$ & 32.08 & 2.606 & 0.4177 & 0.2294 \\ \midrule
       \textbf{Ours} & \shadowstroke~& synthetic & $\checkmark$ & \textbf{32.41} & \textbf{3.006} & \textbf{0.4441} & \textbf{0.2373} \\
    \bottomrule
    \end{tabular}
    }
    \vspace{-2mm}
    \captionsetup{font=small}
    \caption{\textbf{Quantitative ablation studies.} The proposed \shadowstroke~conditioning, synthetic training data, and scene parameter optimization all improve the overall generation quality. IR and HPS stands for the ImageReward~\citep{imagereward} and Human Preference Score~\citep{hps}.
    }
    \captionsetup{font=normal}
    \label{tab:ablation}
\end{table*}

\subsection{Applications}
\label{sec:application}

We demonstrate the versatility of our framework through four applications: (i) generating diverse shadow-drawing art from a single object, (ii) creating shadow-drawing art from multiple input objects, (iii) extending to animated objects, and (iv) deploying in real-world setups.

\smallskip
\noindent\textbf{One Object, Diverse Results.} 
Our framework by design generates multiple shadow–drawing compositions from a single object. By varying the light direction, object pose, and line drawing, we obtain a set of artworks that highlight different structural aspects of the same object. This demonstrates how a single physical object can serve as the basis for a broad range of artistic expressions. Qualitative examples are shown in Fig.~\ref{fig:application}(c). As discussed in Sec.~\ref{sec:scene-config-select}, our framework also supports user-specified subjects through prompt control; see supplementary~\ref{appendix:user} for details.

\smallskip
\noindent\textbf{Multi-object Compositions.}  
Our framework naturally extends to scenes involving multiple objects. For each candidate configuration, we independently sample self-rotation angles, arrange the objects vertically, and release them in Blender’s physics simulation to obtain a stable stacked layout. Once equilibrium is reached, the configuration is treated as a single composite object, allowing the rest of the pipeline to be applied directly. This enables more elaborate results where different objects contribute complementary shadow structures. See Fig.~\ref{fig:application}(b) for examples.

\smallskip
\noindent\textbf{Animated Shadow-drawing Art.}  
Our framework also supports animated objects without extra training. For each configuration, we render five key frames and overlay their \shadowstrokes into a single image, using distinct colors to denote frames. This composite is then fed to the VLM to generate the corresponding prompt. As in the static-object setting, we apply a binary mask to restrict stroke placement, defined as the intersection of all shadow regions and their neighborhoods, to avoid strokes in dynamically changing areas (details in supplementary~\ref{appendix:animated}). We evaluate this pipeline on animated objects from Objaverse, with qualitative results in Fig.~\ref{fig:application}(d) demonstrating the ability of our system to handle temporally varying scenes.

\smallskip
\noindent\textbf{Real-world Deployment.}  
Our framework can be readily reproduced in physical settings without any specialized equipment, requiring only an object and a single spotlight. In practice, everyday household items combined with a phone flashlight suffice to create compelling shadow–drawing compositions. This accessibility positions our approach as a practical tool for artists and hobbyists, significantly lowering the barrier to exploring computational shadow art. Fig.~\ref{fig:teaser} presents physical prototypes of the letters C, V, P, R, with a complete end-to-end demonstration provided on the project page in the supplementary material.

\vspace{-2mm}
\section{Conclusion}
\label{sec:conclusion}

We introduce \textsc{ShadowDraw}, a framework for creating unified shadow–drawing compositional art from arbitrary 3D objects. Our approach optimizes scene parameters to discover semantically meaningful configurations, employs \shadowstrokes to guide line drawing generation, and incorporates automatic evaluation and ranking to ensure shadow-drawing coherence and visual quality. Experiments show strong results across diverse 3D assets, with natural extensions to multi-object scenes, animations, and real-world setups. By broadening the design space of computational visual art, our work opens new avenues for accessible, democratized creation of shadow-based art.

\noindent\textbf{Limitations.}  
While our framework consistently generates compelling results across diverse objects, certain challenges remain. Some objects naturally produce uninformative shadows, making it hard to form high-quality compositions. The diffusion inference adds runtime overhead, and although automated ranking surfaces strong candidates, occasional human judgment is still needed to find the best outcome. We provide more discussions in the supplementary~\ref{appendix:limitations}.

 \IfDefinedSwitch{\arxiv}
{\section*{Acknowledgment}
The research is partially supported by a gift from Ai2, NVIDIA Academic Grant, and DARPA TIAMAT program No. HR00112490422. Its contents are solely the responsibility of the authors and do not necessarily represent the official views of DARPA.}
{}



\bibliography{ref/ref}
\bibliographystyle{ref/ref}

\newpage
\clearpage
\setcounter{page}{1}
\setcounter{section}{0}
\maketitlesupplementary

\section{Supplementary Material Overview}
In this supplementary material, we provide additional implementation details and present extended qualitative results. 
\IfDefinedSwitch{\arxiv}
{Check out our~\projectpage~for more results and an end-to-end real-world demonstration of our pipeline!}
{For more results and an end-to-end real-world demonstration of our pipeline, please refer to the \textbf{project page} in the supplementary document.}

\section{Implementation Details}
\label{appendix:impl-details}

\subsection{Training Data Construction}
\label{appendix:data-contruction}
We construct a paired dataset of \shadowstrokes~and line drawings to train our line drawing generation model. The pipeline proceeds as follows. We first generate 100 line drawings using GPT-4o, prompting it with descriptions of everyday objects. We retain only those drawings that contain at least one closed region bounded by strokes. Next, we fine-tune FLUX-1-dev LoRA~\citep{flux} on this filtered subset and employ it to synthesize an additional 10K line drawings from GPT-4o-generated prompts describing everyday subjects. Finally, we apply OpenCV’s \texttt{FindContour} algorithm to extract closed regions from the synthesized drawings and use a greedy merging strategy that iteratively combines the two smallest connected regions until only four remain. The closed contours of these merged regions (and their union) serve as the \shadowstroke~conditions for training.

\begin{figure}[h]
    \centering
    \includegraphics[width=\linewidth]{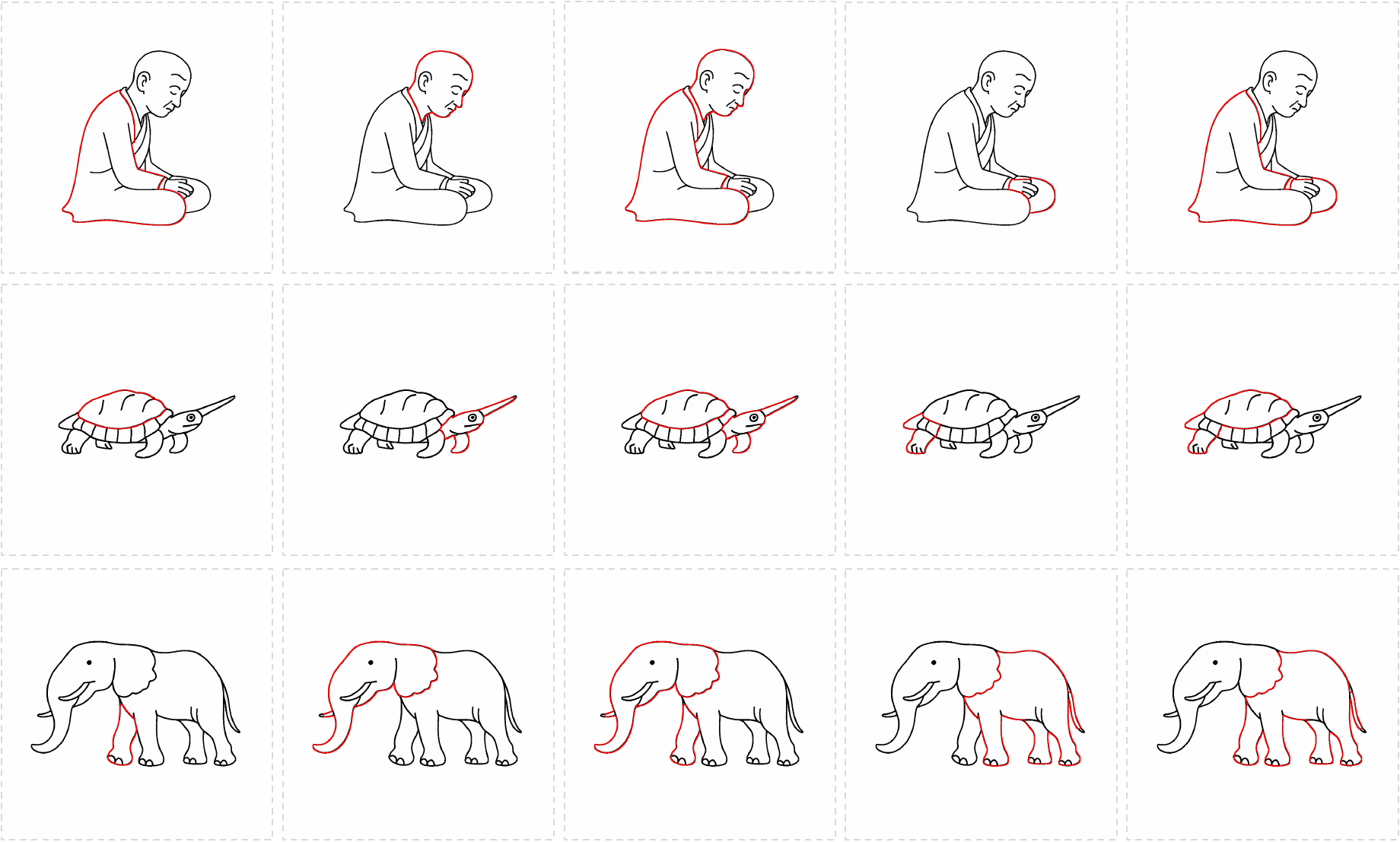}
    \captionsetup{font=small}
    \caption{Examples of training data pairs. Each row shows a line drawing generated by our finetuned FLUX model, with different closed contours extracted from it. Each image forms a training pair, where the red contour is used as the condition and the full line drawing serves as the target.}
    \captionsetup{font=normal}
    \label{fig:placeholder}
\end{figure}

\subsection{Visual Prompt Proposal} 
\label{appendix:visual-prompt-proposal}
Vision–language models are highly sensitive to input formatting, and poorly structured inputs often result in uninformative or inconsistent outputs. To address this, we carefully design a system prompt that guides the VLM, specifically GPT-4.1, to generate detailed, semantically meaningful descriptions of the provided stroke and its role in the complete line drawing. The full system prompt is given below.

\begin{lstlisting}[language={},caption={System prompt for creating the textual description of the intended line drawing in the visual prompt proposal stage.}]
You are a skilled artist specializing in expressive, imaginative, and visually striking minimalist line drawings. You will be shown an image containing a contour. Your task is to interpret this contour and create a complete line drawing, using the provided contour as the core expressive element of your composition. The subject you draw should be a character, either a human, an animal, or a cartoon or anthropomorphic figure.  

# Instructions  

First, analyze the contour to identify its function in the drawing. Follow these steps:  

1. Analyze the contour's geometry and position on the canvas.  

2. Determine the subject of the line drawing and which major, prominent body part (such as body, head, face) or clothing (such as skirt or dress) the contour outlines. Do not describe small or less essential features, such as hands, tails, wings, or beaks. Specify the subject in precise terms:  
For people, use identifiers (e.g., man, woman) or vocations (e.g., dancer, sailor, guitarist);  
For animals, name the species (e.g., bird, fish, cat, dog);  
For cartoon or anthropomorphic characters, name the type (e.g., ghost, robot, cookie character, book character).  

3. Explain your reasoning in detail, including the stroke's shape, its position on the canvas, and why it is a good fit for the composition.  

Next, write a description of the complete drawing without referencing the provided contour, following this structure:  

1. Opening: A minimalist line drawing of a [character] [in a pose or with an expression], matching your earlier interpretation.  

2. Physical description: The [character] has [facial feature or expression] and wears [clothing or accessories].  

3. Object or motion (optional): The [character] is [doing something, holding something, or in motion].  

4. Gesture or interaction (optional): Further describe the subject's gesture or interaction with their environment.  

5. Conclude with a style remark: The style is [adjective(s)], [additional notes about technique or focus].  

# Format requirement  

1. Separate the two parts with a blank line.  

2. Do not use numbering, bullet points, or extra formatting.  

3. Strictly follow this structure without additional comments:  

The provided contour shows an outline of the [specific body part or clothing] of a [character]. The reason is [contour geometry interpretation]. [Additional reasons for your interpretation].  

A minimalist line drawing of a [character] [in a pose or with an expression]. The [character] has [facial feature or expression] and wears [clothing/accessories]. [Optional action or motion]. [Optional gesture or interaction]. [Artistic style remark].
\end{lstlisting}

\subsection{Baseline Execution Details}
\label{appendix:baseline-execution}

Here we describe how we use Gemini Flash 2.5 Image (a.k.a.\ nano banana)~\citep{gemini} to generate shadow–drawing art. In the \emph{object–shadow} version, we provide the model with the object–shadow composite and the line drawing description produced by our approach, and ask it to directly generate a shadow–drawing composition. In the \emph{\shadowstroke} version, we instead provide the \shadowstroke~and the line drawing description, prompting the model to complete the drawing. As in our framework, we then remove the input \shadowstroke~from the generated drawing, reinsert the 3D object, and render the final composition, where the cast shadow completes the drawing.

\subsection{Discussion on the Evaluation Algorithm}
\label{appendix:ranking}

\begin{figure*}[h]
    \centering
    \includegraphics[width=\linewidth]{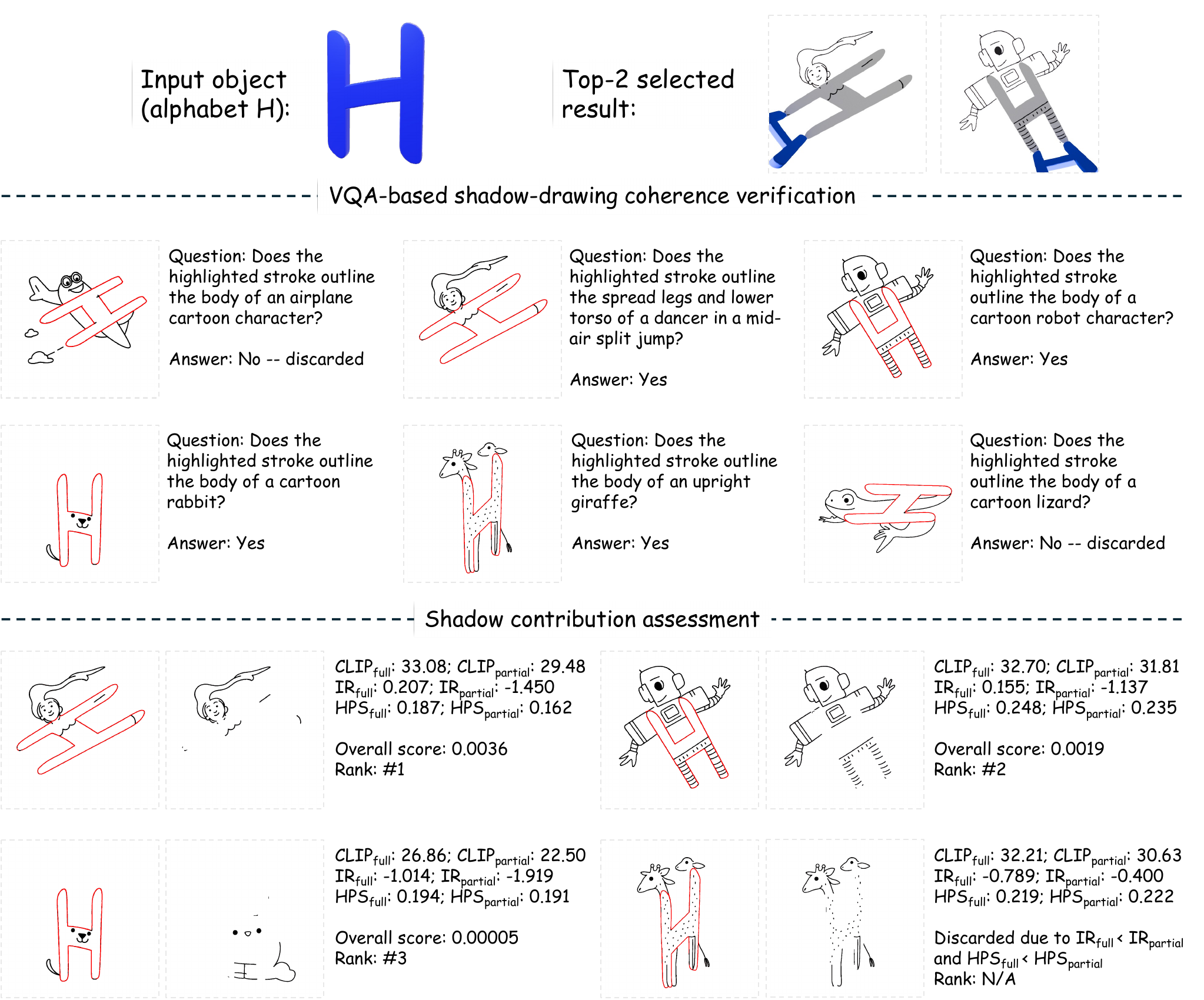}
    \caption{Illustration of the evaluation and ranking process, which discards incoherent cases and preserves only those where the shadow meaningfully contributes to the composition.}
    \vspace{-2mm}
    \label{fig:ranking}
\end{figure*}

\noindent\textbf{Visual Illustration.}  
Figure~\ref{fig:ranking} illustrates how our evaluation and ranking algorithm selects high-quality shadow-drawing compositions. In the first stage, the VQA-based verification discards incoherent cases (e.g., when the shadow stroke does not correspond to the intended body part of the character). In the second stage, the shadow contribution assessment compares complete and contour-removed versions, ensuring that the shadow meaningfully enhances the drawing. As shown, this two-step process ranks plausible results higher while filtering out cases where the shadow plays only a minor or misleading role. Overall, the pipeline balances semantic alignment, structural coherence, and visual quality, producing consistent and interpretable rankings.

\noindent\textbf{Analyses.} Evaluating our generated shadow–drawing compositions is inherently challenging, as their abstract and artistic qualities often resist objective evaluation. To rigorously assess effectiveness, we designed two complementary user studies based on pairwise preference judgments. (1) For each object, we randomly select one result from the top-4 ranked outputs and another from the remaining, asking evaluators to choose their preferred composition; and (2) we randomly select two results among the top-3, where differences are more subtle. In both cases, evaluators may also indicate that neither is preferable. Altogether, we collected 2,000 preference pairs from 10 annotators for the first study, and 2 labels per top-3 pair across all 200 objects for the second study.

In the first study, our ranking algorithm achieves a strong alignment with human judgment, agreeing on 63.5\% of pairs, disagreeing on only 11.0\%, and with 25.5\% of cases marked as no clear preference. These results indicate that our automated ranking provides a reliable proxy for subjective evaluation in the broader design space. In the second study, where all candidates are already of very high quality, the agreement rate with human preference naturally decreases to 39.8\%, with another 24.3\% of cases judged as indeterminate. Crucially, the agreement between two independent sets of human annotations is itself only 44.5\%, underscoring the intrinsic subjectivity of evaluating artistic compositions. Taken together, these findings suggest that while perfect alignment is unattainable in such a subjective domain, our ranking system performs comparably to human consensus and thus provides a practical, scalable tool for curating shadow–drawing art.

\begin{figure*}[t]
    \centering
    \vspace{-2mm}
    \includegraphics[width=\linewidth]{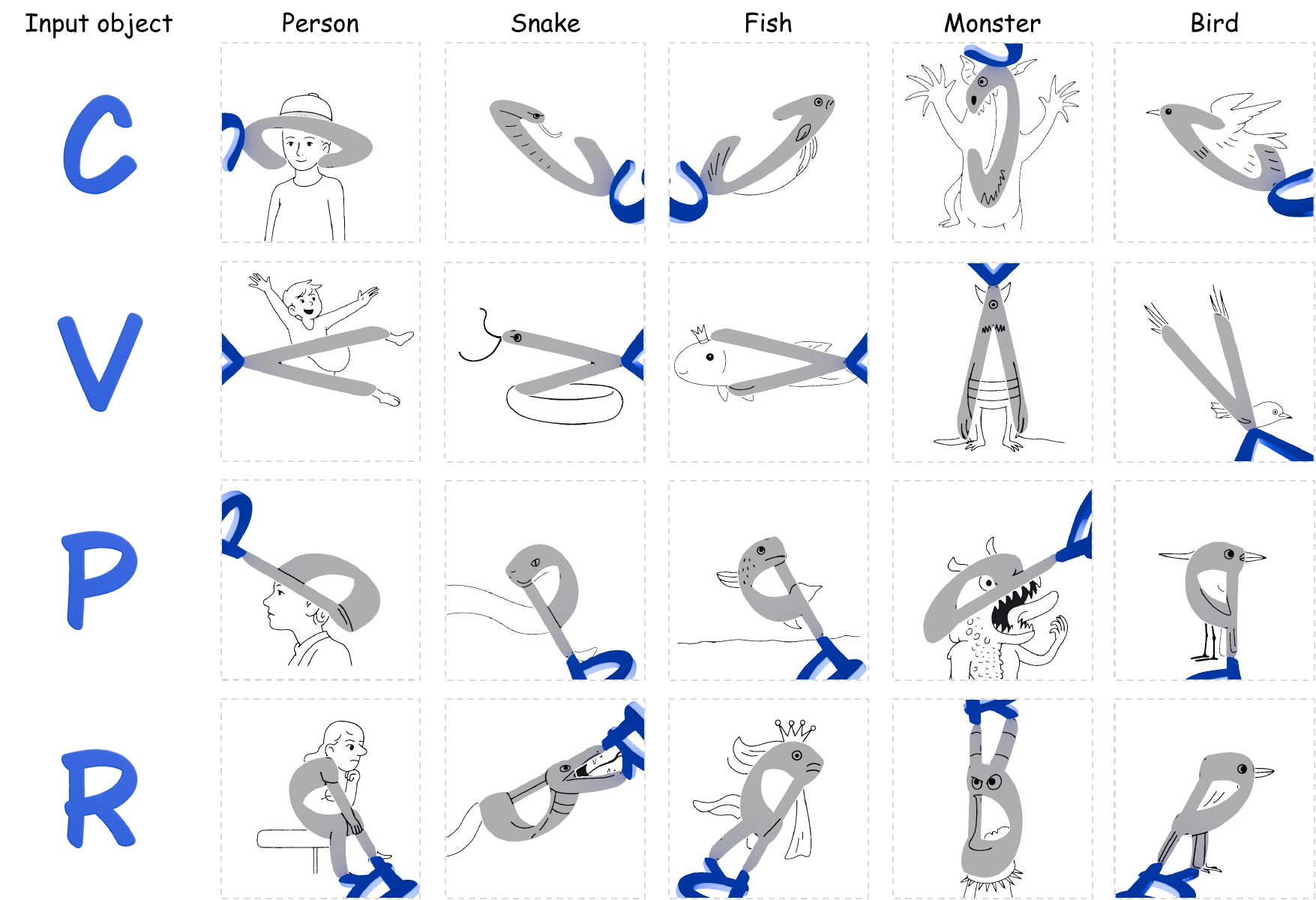}
    \caption{\textbf{Shadow–drawing compositions with user-specified subjects.} 
    We show results for alphabet-shaped objects (C, V, P, R) conditioned on the subjects \emph{ghost}, \emph{fish}, \emph{person}, and \emph{bird}.  
    While the pipeline supports flexible subject control through prompt editing, certain object geometries inherently limit the achievable compositions, leading to occasional unsatisfactory outcomes.}
    \vspace{-1mm}
    \label{fig:specify-subject}
    \vspace{-2mm}
\end{figure*}

\subsection{Subject-Specified Generation}
\label{appendix:user}

We also demonstrate that our pipeline supports user-specified subject control through prompt editing. In practice, the desired subject is directly specified in the system prompt used by the VLM during visual prompt proposal. However, we observe that certain object–subject combinations are inherently incompatible: the geometry of the 3D object may not afford a shadow stroke that can be meaningfully integrated into the specified concept, leading to unsatisfactory results. Examples are shown in Fig.~\ref{fig:specify-subject}.

\subsection{Animated Shadow-drawing Art} 
\label{appendix:animated}

As mentioned in Sec.~4.3, our framework supports animated objects without requiring additional training. We provide the implementation details as follows. For each candidate configuration, we render five keyframes of the animation and extract their \shadowstrokes. To preserve temporal information, we overlay the strokes into a single composite image, assigning distinct colors to each frame so that the VLM can recognize temporal variation and generate a prompt.

A critical step in this pipeline is the construction of a binary mask to restrict where strokes may be placed. Without such a constraint, strokes might appear in regions where shadows change across frames, leading to incoherent results. To build the mask, we proceed as follows. First, we render the shadow silhouettes of all five keyframes. Pixels that are covered by shadows in every frame are designated as the \emph{static region}, while pixels that are covered in at least one but not all frames are designated as the \emph{dynamic region}. Next, for each pixel on the canvas, we compute its distance to both the static and dynamic regions. If a pixel is closer to the dynamic region than to the static region, we mask it out, prohibiting stroke placement in that location. Intuitively, this rule ensures that strokes are confined to stable shadow regions and avoid areas subject to temporal variation.

Once the mask is established, the generation process follows the same procedure as in the static-object setting. The shadow contour and the corresponding VLM-generated prompt are fed to the line drawing generator, and masked regions are excluded during synthesis. Finally, the animated object is reinserted into the scene and rendered across frames, with its shadow complementing the static line drawing to form a temporally consistent shadow–drawing composition.

We evaluate this pipeline using animated objects from Objaverse. Qualitative results are presented in Fig.~6(d) (main paper) and Fig.~\ref{appendix:animated}. As shown, our system successfully generates line drawings that remain coherent with dynamic shadows, demonstrating the ability of our approach to extend from static to temporally varying scenes.

\begin{figure*}[t]
    \centering
    \vspace{-5mm}
    \IfDefinedSwitch{\embedVideo}{\animategraphics[autoplay,loop,controls={play,stop}, width=\linewidth, trim=0 5cm 0 3cm, clip]{12}{figures/animated_object/}{000}{011}
    }
    {\includegraphics[width=\linewidth]{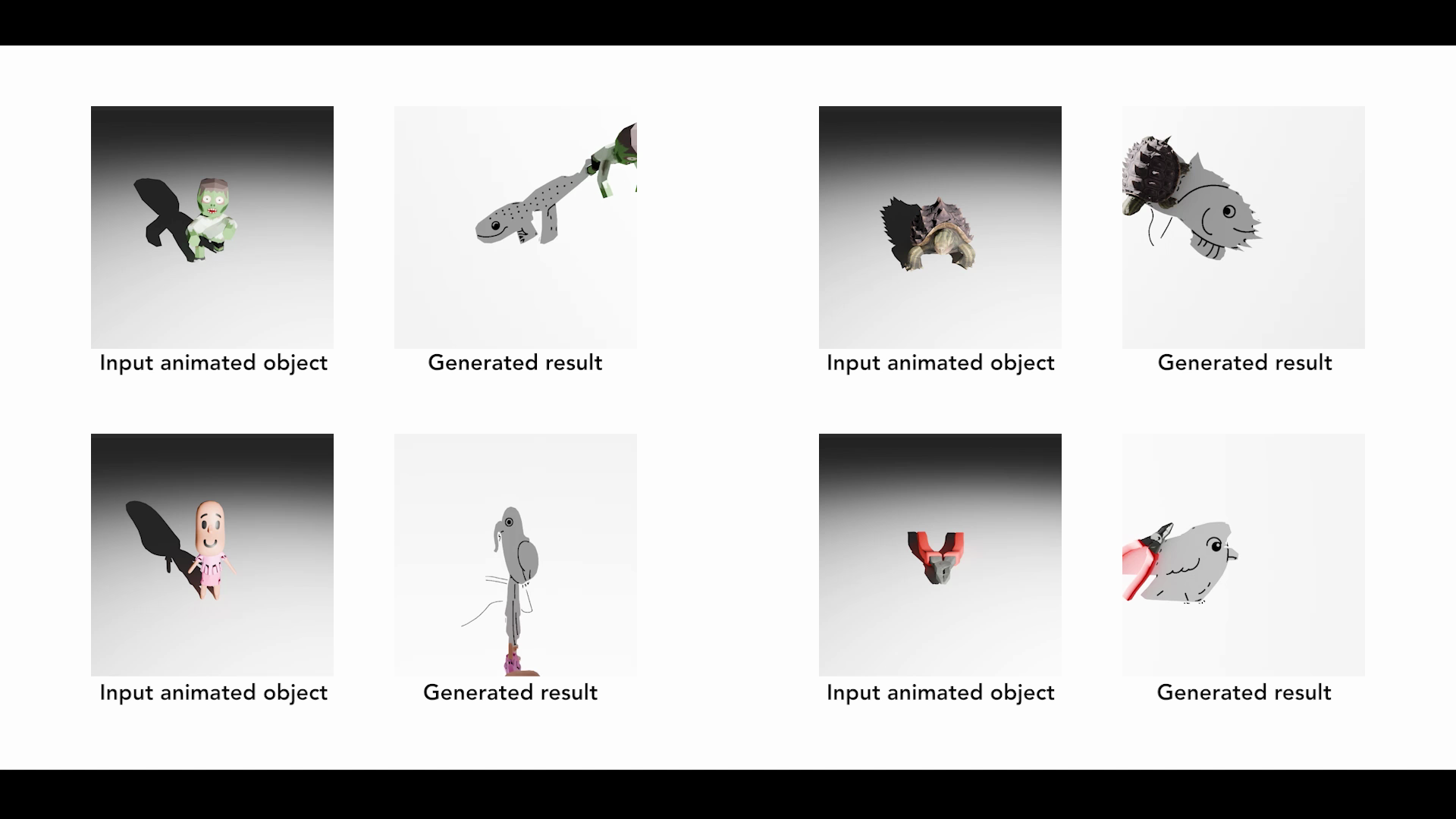}
    }
    \vspace{-2mm}
    \caption{\textbf{Animated shadow–drawing art.} Line drawings generated with our pipeline remain coherent as dynamic shadows complete the composition across frames. \emph{Best viewed in Adobe Acrobat Reader for the embedded animation.}}
    \label{fig:appendix-animated}
    \vspace{-3mm}
\end{figure*}

\begin{figure*}[!t]
    \centering
    \includegraphics[width=0.9\linewidth]{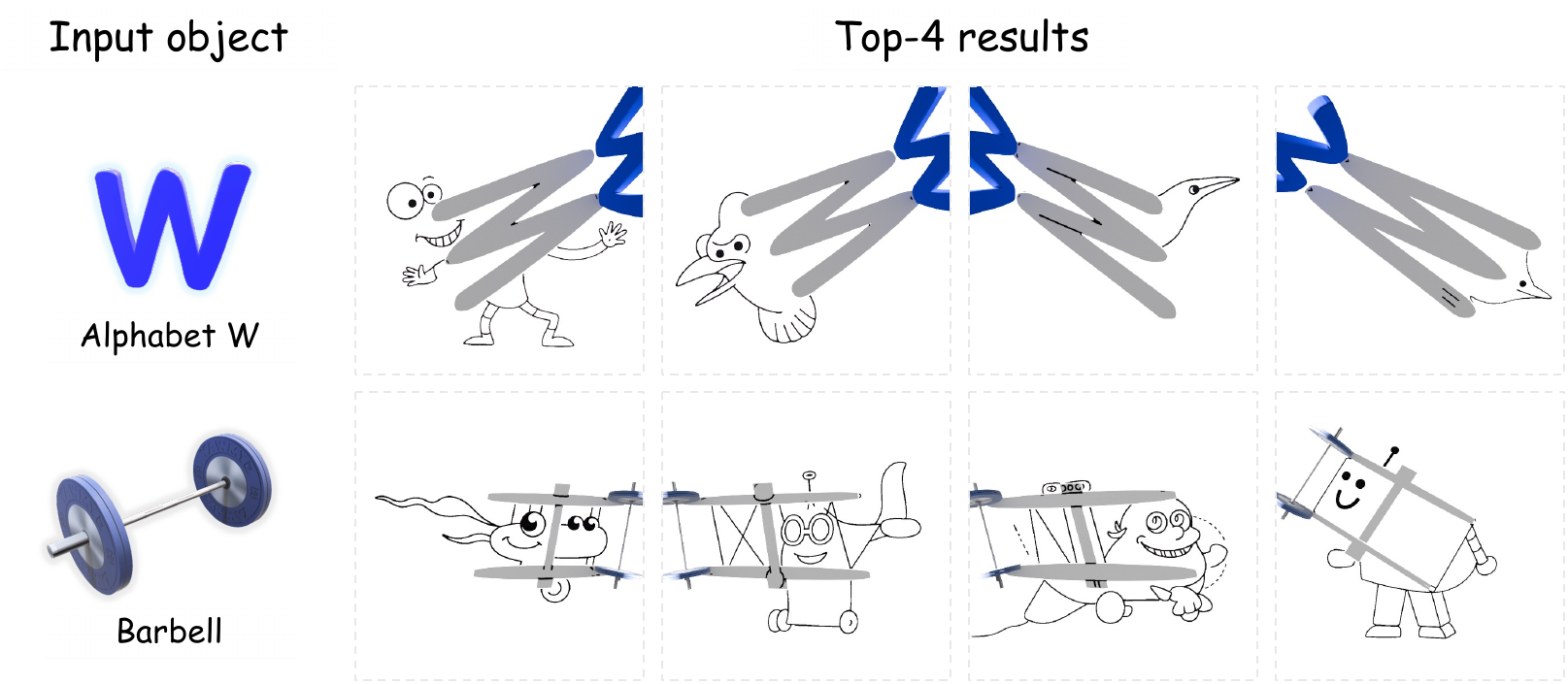}
    \caption{\textbf{Failure cases.} Some objects produce shadows that are ambiguous or uninformative, making it difficult for our system to produce meaningful shadow-drawing compositions.}
    \label{fig:failure-objects}
\end{figure*}

\subsection{Runtime Analysis}
\label{appendix:runtime}

The only trainable component in our framework is the line drawing generation model based on FLUX.1-Canny~\cite{flux}. Specifically, we train a DoRA~\cite{dora} adapter on all queries, keys, values, and MLPs of the backbone diffusion transformer. We use the Adam optimizer with a constant learning rate of $10^{-4}$ and train for roughly 12 hours on 8~A6000 GPUs. At inference, the dominant cost arises from diffusion sampling, taking about 40~seconds per image with 30 steps. For a single object, generating line drawings for 48 sampled scene configurations requires approximately 30~minutes, and the full pipeline completes in about 35~minutes on a single A6000 GPU. Reducing the number of inference steps from 30 to 10 lowers latency to around 15~minutes with minimal quality degradation. Because the process is fully parallelizable, latency can be reduced to under 5~minutes on standard 8-GPU nodes. Further acceleration may be achieved by distilling the multi-step diffusion process into a one- or few-step generator.


\section{Limitations}
\label{appendix:limitations}

While our method enables diverse and visually engaging shadow–drawing compositions, several limitations remain. First, the quality of results is closely tied to the intrinsic shape of the object: some objects inherently produce shadows that are either visually uninteresting or too ambiguous to interpret, regardless of lighting or pose. As illustrated in Fig.~\ref{fig:failure-objects}, such cases often yield shadows that lack recognizable structure or fail to align meaningfully with the generated drawing. Second, the joint search and generation process over scene parameters introduces noticeable runtime overhead. Although this procedure is necessary to explore the large design space, generating results for a single object still takes a relatively long duration. Finally, while our ranking algorithm is generally effective at surfacing strong candidates, it is not flawless. In practice, users may still need to examine multiple outputs to identify the most compelling result. Addressing these limitations through richer shadow descriptors other than fractal dimension, more efficient search strategies, and refined ranking or user-in-the-loop mechanisms represents promising directions for future work.

\end{document}